%% file: main.tex
\documentclass[letterpaper]{article} 
\newif\ifanonymous
\anonymousfalse 
\ifdefined\arxivbuild\anonymousfalse\fi 
\ifanonymous
  \usepackage[submission]{aaai2027}
\else
  \usepackage{aaai2027}
\fi
\ifanonymous\else\nocopyright\fi 
\ifanonymous\else\fi 
\usepackage[hyphens]{url}  
\usepackage{graphicx} 
\usepackage{natbib}  
\usepackage{caption} 
\usepackage{booktabs}
\usepackage{amsmath,amssymb}
\usepackage{enumitem}
\usepackage{xcolor}
\definecolor{reglosblue}{HTML}{2F5D8A} 
\definecolor{reglostint}{HTML}{EEF4FB}
\usepackage[most]{tcolorbox}
\newtcolorbox{promptbox}[1]{enhanced, breakable, colback=reglostint, colframe=reglosblue, boxrule=0.6pt, arc=2pt,
  left=4pt, right=4pt, top=3pt, bottom=3pt, fonttitle=\footnotesize\bfseries\sffamily, coltitle=white,
  title={#1}, fontupper=\footnotesize}      
\title{Reglos: A Self-Evolving Personal Memory Agent\\with Recursive Self-Improvement}
\author{
    Sen Yang\textsuperscript{\rm 1}\equalcontrib,
    Boqin Yuan\textsuperscript{\rm 2}\equalcontrib,
    Jiaduo Lin\textsuperscript{\rm 1},
    Zehong Huang\textsuperscript{\rm 1},
    Wen Liu\textsuperscript{\rm 1}\corresponding
}
\affiliations{
    \textsuperscript{\rm 1}Reglos AI \qquad \textsuperscript{\rm 2}UC San Diego\\
    \texttt{\{sen, jiaduo.lin, zehong.huang, wen\}@reglos.ai} \quad \texttt{b4yuan@ucsd.edu}
}

\usepackage{etoolbox}
\begin{document}
\makeatletter\patchcmd{\@maketitle}{These authors contributed equally.}{Equal contribution.}{}{\PackageWarning{main}{equal patch failed}}\patchcmd{\@maketitle}{These authors contributed equally.}{Equal contribution.}{}{\PackageWarning{main}{equal patch failed}}\makeatother 
\maketitle

\begin{abstract}
\input{sections/0_abstract}
\end{abstract}

\input{sections/1_introduction}
\input{sections/2_related_work}
\input{sections/3_reglos}
\input{sections/6_results}

\input{sections/7_discussion}
\input{sections/8_conclusion}

\bibliography{references}

\appendix
\ifanonymous\else\renewcommand{\arraystretch}{0.9}\setlength\floatsep{7pt plus 2pt}\setlength\textfloatsep{11pt plus 2pt}\fi 
\input{sections/A_impl}
\input{sections/A_variants}
\input{sections/A_run}
\end{document}

%% file: sections/0_abstract.tex
Personal agents draw on past conversations to tailor their responses to each user. Their memory systems store and update what each user says, and coding agents now also rewrite the harness that turns this memory into the model's context. Such patches risk overfitting, because one tested only on the failures it targets can fix those cases and help nowhere else. We present Reglos, a personal memory agent that tests whether harness changes help beyond the failures they were written to fix. Its self-evolving memory adds to each user's append-only record with every conversation, as new preferences supersede old ones and facts can be forgotten. Its harness improves through recursive self-improvement, in which a coding agent proposes patches from failure traces and a critic from a second model family screens each one for benchmark gaming. After the harness is frozen, each further patch is tested on users the loop has not seen. With the same backbone behind every system, Reglos outperforms state-of-the-art memory systems such as Mem0, MemOS and EverOS, reaching 67.7\% and 84.0\% on PersonaMem-v2 and LongMemEval-S with Qwen3.8-27B, and 62.1\% and 87.2\% with GPT-5.6 Luna.

\ifanonymous\else 
\medskip
\centerline{\raisebox{-0.22em}{\includegraphics[height=1.15em]{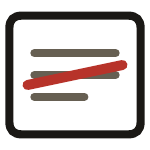}}\ \ \textbf{Website:}\ \textcolor{reglosblue}{\url{https://reglos.ai}}}
\fi

%% file: sections/1_introduction.tex
\section{Introduction}
\label{sec:intro}
A personal assistant that talks with the same user for months has to keep track of what that user prefers \citep{hu2025memorysurvey,jiang2025personamemv2}. Some of those preferences change, and the user may later ask the assistant to forget others. Reading the whole history handles this poorly, and even frontier models often miss a preference the user only implied or keep applying one that has changed \citep{jiang2025personamemv2}. Memory systems keep what each user says across conversations and recall it when the model answers \citep{packer2023memgpt,chhikara2025mem0,rasmussen2025zep,li2025memos,hu2026evermemos}. How well they work depends on what they store and on the code that decides what to recall and how to show it to the model, which we call the \emph{harness}.

\begin{figure}[t]
\centering
\includegraphics[width=\columnwidth]{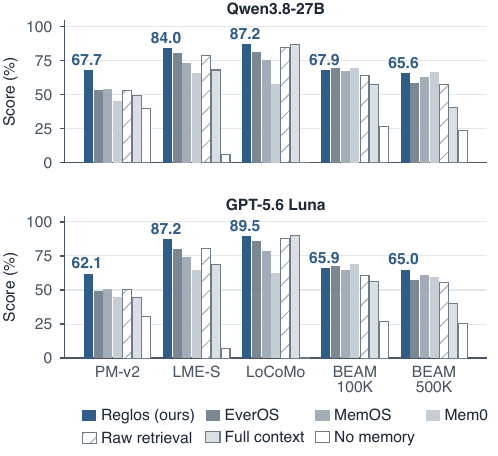}
\caption{Qwen3.8-27B writes memory for every system and answers in the top panel; GPT-5.6 Luna answers in the bottom panel. Reglos leads on PersonaMem-v2 and LongMemEval-S. Details in Table~\ref{tab:main}.}
\label{fig:results}
\end{figure}
\begin{figure*}[t]
\centering
\includegraphics[width=\textwidth]{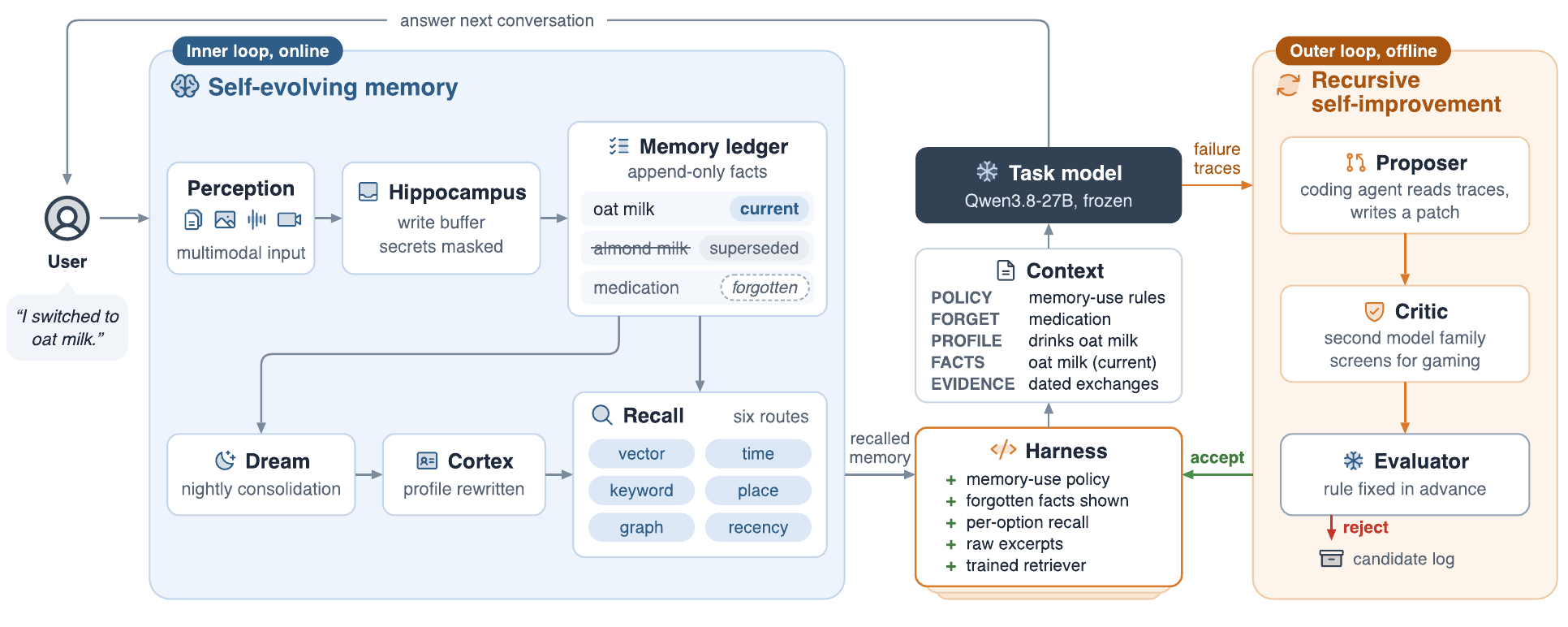}
\caption{Reglos improves on two time scales: each user's memory evolves online (inner loop), and the harness that turns memory into context for a frozen model is recursively improved offline (outer loop).}
\label{fig:overview}
\end{figure*}
Although these systems update what they store as the user talks, the harness itself is usually written by hand and revised between releases. Recent work hands that job to coding agents, which evolve memory programs and pipelines against benchmarks \citep{liu2026mempro,chen2026prismem,zhang2025memevolve}. A benchmark score alone leaves one question open: does a change help users whose failures nobody inspected while writing it?

Our own development data show why that question matters. A memory can be stored correctly and still be used wrongly: in 31\% of the failures we traced, the evidence the model needed was already in its context (Section~\ref{sec:res-pm}). Those are failures of the harness, and a coding agent can patch them, but a patch judged on the cases that prompted it can look better than it is.

These failures point to two kinds of change with different owners. A user's memory changes whenever that user talks, and an error in it is best fixed then, inside that user's record. The harness serves every user at once, so a change to it can help some users and hurt others, and that is where overfitting happens. We therefore argue that a personal memory system should improve at two time scales: its memory online as each user talks, and its harness offline through changes that a fixed evaluator accepts and that are then tested on users the loop has not seen.

We present \textbf{Reglos}, a personal memory agent built on this principle (Figure~\ref{fig:overview}). Its inner loop is a self-evolving memory. Each fact enters an append-only record that stores when it holds and when it was learned, so a new preference can supersede an old one without erasing it. A request to forget expires the matching facts, and a nightly pass rewrites the profile from the user's current facts. Because these states are explicit, the harness can tell the model which of two conflicting preferences is current and what it must not use. Its outer loop improves the harness through recursive self-improvement. A coding agent reads failure traces and writes patches, and a critic from a second model family clusters the same failures and rejects patches that game the benchmark. After the harness is frozen, each further patch is tested on users the loop did not see during development. The two loops co-evolve. The outer loop learns from the failures of the memory the inner loop builds, and each accepted patch changes how that memory is written and read. Qwen3.8-27B writes memory throughout.

With Qwen3.8-27B writing memory for every system, Reglos outperforms Mem0, MemOS and EverOS on PersonaMem-v2, LongMemEval-S and LoCoMo, whether Qwen3.8-27B or GPT-5.6 Luna answers (Figure~\ref{fig:results}). On the official PersonaMem-v2 benchmark split it reaches 67.7\%, 13.9 points above the strongest of them; published results reach 57.2\% with a GPT-4.1-mini reader \citep{liu2026d2acci} and 59.3\% with kimi-2.5 \citep{fei2026dcpm}. On LongMemEval-S it reaches 84.0\%, 3.6 points above EverOS. With GPT-5.6 Luna answering, it scores 62.1\% and 87.2\% and again leads both benchmarks. Within the outer loop, the coding agent's accepted patches lift the harness from 54.2\% to 60.7\% on the validation questions, and a later patch that adds a trained retriever raises the test-set result from 63.4\% to 67.7\%.

Our contributions are:
\begin{itemize}[leftmargin=*]
\item \textbf{A self-evolving personal memory agent.} Reglos keeps an append-only memory for each user that records how preferences change and what the user asked to forget (Section~\ref{sec:reglos}).
\item \textbf{Recursive self-improvement of the harness.} A coding agent proposes patches from failure traces, a critic from a second model family screens them, and after the freeze each further patch is tested on users the loop did not see (Sections~\ref{sec:loops} and~\ref{sec:res-loop}).
\item \textbf{Results across backbones.} Reglos outperforms Mem0, MemOS and EverOS on three benchmarks under two answering models and exceeds the best published result on PersonaMem-v2 (Section~\ref{sec:results}).
\end{itemize}

%% file: sections/2_related_work.tex
\section{Related Work}
\label{sec:related}

\paragraph{Memory Systems and Benchmarks.}
Early memory for LLM agents paged information in and out of a bounded context \citep{packer2023memgpt} or kept a stream of observations with periodic reflection \citep{park2023generative}. Production systems now extract and reconcile facts \citep{chhikara2025mem0}, build temporal knowledge graphs \citep{rasmussen2025zep}, link notes \citep{xu2025amem} or manage memory as an operating system \citep{li2025memos,hu2026evermemos}; \citet{hu2025memorysurvey} survey the space. Benchmarks have moved from recalling facts in long dialogues \citep{maharana2024locomo,wu2025longmemeval} to inferring implicit, evolving preferences \citep{jiang2025personamem,jiang2025personamemv2} and to long agent trajectories \citep{zhao2026amabench}. Two diagnostic studies shape our evaluation. Most memory systems fall below naive retrieval over the raw history on PersonaMem-v2 \citep{he2026memorytax}, so we report raw retrieval with our own answer policy as a control. On LoCoMo, most failures arise at retrieval (Yuan et al.~\citeyear{yuan2026diagnosing}); on PersonaMem-v2 we find most reachable failures after retrieval, when the answering model already holds the evidence (Section~\ref{sec:results}).

\paragraph{Self-Evolving Agents and Memory.}
Self-evolving agents improve from their own interaction data and feedback \citep{gao2025selfevolving,fang2025selfevolving,zhang2026metaagentx}; without retraining the underlying model, they reflect on failures \citep{shinn2023reflexion,madaan2023selfrefine}, accumulate skills \citep{wang2024voyager}, or grow an evolving context or prompt \citep{zhang2025ace,agrawal2025gepa}. Memory is a common place where this evolution happens. MemEvolve meta-evolves how experience is encoded, stored and retrieved \citep{zhang2025memevolve}, EvolveMem tunes retrieval configurations from failure logs \citep{liu2026evolvemem}, MemSkill and SelfMem evolve memory skills and strategies \citep{zhang2026memskill,yang2026selfmem}, and MindMemOS evolves memory schemas and consolidates them by dreaming \citep{liang2026mindmemos}; Memory-R1, Mem-$\alpha$ and Mem-T instead train the memory manager with reinforcement learning \citep{yan2025memoryr1,wang2025memalpha,yue2026memt}. Reglos's inner loop is self-evolution of this kind, applied per user under fixed rules (supersession that keeps history, retraction that records the request, and user edits).

\paragraph{Recursive Self-Improvement.}
In the sense of \citet{duan2026lastai}, recursive self-improvement (RSI) is an autonomous, closed loop whose changes persist and alter later improvement. One line of recent work trains the model inside that loop \citep{lin2026lmimprove,zhang2026rsipo}. Another is training-free and evolves the harness around a frozen model \citep{lee2026metaharness,zhang2026selfharness,lee2026rhi}, and some systems combine the two \citep{wei2026selfhpo,su2026coevorsi}. RSI has also reached memory: MemPro, PrisMem and ALMA evolve the memory pipeline as code \citep{liu2026mempro,chen2026prismem,xiong2026alma}. D$^2$ACCI pairs the inference-time memory loop with an outer loop that labels each regression with the pipeline stage that caused it and accepts, flags or rolls back each memory feature through paired statistical tests and checks on protected slices \citep{liu2026d2acci}. Self-Improving Memory has the task model itself rewrite, from its task trajectories, the two prompts that set what memory it loads before a task and what it stores afterwards, and then distills the trajectories run under the rewritten prompts into its weights by fine-tuning \citep{chen2026selfimprovingmemory}. Reglos belongs to the harness line, at the level of strategy autonomy: a coding agent proposes harness code over each user's append-only record, a critic from a second model family reviews every patch, the task model stays frozen, and the accepted harness is evaluated on users the loop has not seen.

%% file: sections/3_reglos.tex
\section{Reglos}
\label{sec:reglos}

Reglos pairs an online loop that updates each user's memory with an offline loop that recursively improves the harness supplying context to the frozen model, and the two co-evolve (Figure~\ref{fig:overview}). The same memory service serves our own app, and Appendix~\ref{app:impl} gives its thresholds and limits.

\subsection{Inner Loop: Self-Evolving Memory}
\label{sec:inner}

\paragraph{From conversation to facts.}
Each user's memory grows from what that user says. New messages wait in a buffer until there are enough to be worth reading, and the task model then extracts the facts, commitments and requests to forget that they contain. Before any of this text reaches a model, secrets such as keys and card numbers are masked. Once extracted, each candidate is compared with the facts already stored. One batch call then decides whether it is new, already known, a replacement for an older fact, or extra detail for one. The accepted facts are kept in the language the user spoke and indexed for recall.

\begin{figure}[!ht]
\centering
\includegraphics[width=\columnwidth]{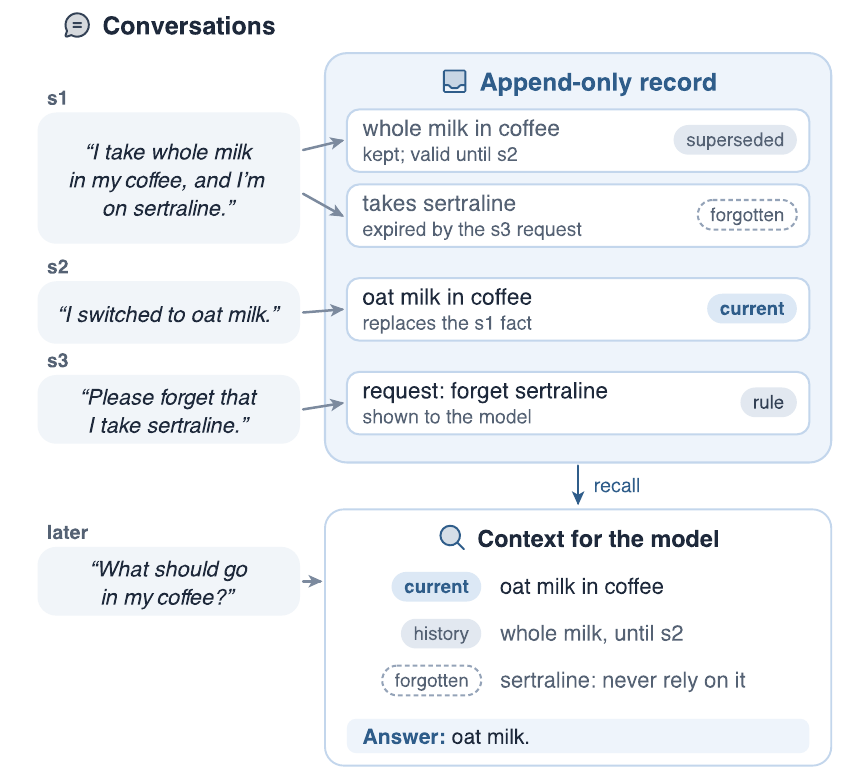}
\caption{One user's record across three sessions. A change of mind closes the old fact and keeps it, a request to forget expires the fact and stays as a rule, and recall shows the model which version holds.}
\label{fig:ledger}
\end{figure}

\paragraph{An append-only record.}
Every fact keeps the identifiers of its source messages, a perspective (the user's own belief, an objective fact, or something the assistant said) and two times: when the fact holds and when the system learned it. A fact's text is never rewritten, so a change of mind closes the old fact's validity and leaves it in the record, where the history of a preference stays readable. A request to forget expires the matching facts and records the request itself, which the harness later shows the model with the rule never to rely on it. Figure~\ref{fig:ledger} follows one user through these cases. Users can also inspect, correct, delete and undo any memory in the app.

\paragraph{Consolidation and recall.}
The record is condensed as well as kept. Each night a pass called Dream summarizes the day's sessions and rewrites the user's profile from the facts that are still current. When a question arrives, recall searches the facts through six routes at once: three on content (vector, keyword and entity graph), two anchored on the time and place the question mentions, and recency. The content routes decide what can come back. Time, place and recency only raise facts that a content route has found, unless no content route finds anything, and a fact found by recency alone is dropped. The routes are fused by reciprocal rank fusion into a fixed token budget, and an \emph{explain} call reports which route surfaced each item.

\subsection{Engine and Serving}
\label{sec:engine}

\paragraph{A Rust kernel and a Python harness.}
The memory kernel is a Rust service. The harness, the benchmark drivers and the outer loop are written in Python and call it over HTTP. The split sets what a candidate costs the outer loop. A harness change is scored on the memories already stored, while an engine change must re-ingest every history.

\paragraph{One database.}
The kernel keeps every layer of memory, and the work queue that maintains it, in a single PostgreSQL~16 database. The indexes behind recall's three content routes live there too, as embedding columns beside the facts, a trigram index and a relational graph table, so no separate vector store is needed. Recall scans each user's rows exactly. One user holds a few hundred to a few thousand facts, and an approximate index filtered down to one user can silently return fewer of them than asked. Extraction and Dream run as jobs that workers lease from the same database, so several instances share the load and a job interrupted by a crash is picked up when its lease expires.

\paragraph{Writes that never lose a fact.}
A candidate that matches a stored fact after normalization (case, width, spacing and punctuation) only confirms the stored row, with no model call. If the batch adjudication fails, every candidate is kept as new, so a model error can duplicate a fact but never lose one. Model calls back off for about three minutes on rate limits and server errors. With the model's reasoning disabled, the engine uses a quarter of the output tokens and 2.6 seconds per model call against 9.3.

\subsection{The Harness}
\label{sec:harness}

The harness is the program that sits between the memory kernel and the frozen task model and determines what the model reads. The kernel supplies the mechanisms, and the harness decides the policy for using them. The harness is the layer the outer loop changes. It receives the question, together with its answer options where the benchmark provides them, but has no access to the dataset's labels. Memory is reachable only through a narrow interface. It offers recall, the list of stored items, the user's profile, the source messages of each fact, and the user's raw conversation, both as excerpts with full-text search and as exchanges (a user message and the assistant's reply) that a retriever can rank (Section~\ref{sec:loops}). The harness can also call the model that writes memory and keep a derived text layer for each user. Within this interface the harness may issue several recall queries, filter and rank the results across memory layers, and decide how the evidence is organized. These decisions form the search space of the outer loop (Section~\ref{sec:loops}).

Each context opens with a memory-use policy and the user's requests to forget, followed by the profile, the recalled facts with superseded ones marked as no longer true, and a small number of session summaries and raw excerpts. For multiple-choice questions the harness also retrieves evidence for each option. The options of a question typically share a template, so recall on the full option text returns nearly the same facts for all of them; the harness therefore queries with the terms that distinguish each option from the others. This per-option retrieval was the final step accepted by the outer loop (Section~\ref{sec:res-loop}).

The policy specifies how the model should interpret the states that the record encodes. The model must not rely on facts the user asked to forget or on preferences that are no longer current, and it must keep the preferences of other people separate from the user's own. Further rules address what the record cannot encode: the memory is incomplete, preferences must not be inferred from demographic attributes, and sensitive identifiers must not appear in the answer. Appendix~\ref{app:impl} gives the full policy.

\subsection{Outer Loop: Recursive Self-Improvement}
\label{sec:loops}

\begin{table*}[t]
\centering
\small
\setlength{\tabcolsep}{4.6pt}
\begin{tabular}{@{}l ccccc ccccc@{}}
\toprule
 & \multicolumn{5}{c}{\textbf{Qwen3.8-27B}} & \multicolumn{5}{c}{\textbf{GPT-5.6 Luna}} \\
\cmidrule(lr){2-6}\cmidrule(l){7-11}
& & & & \multicolumn{2}{c}{BEAM} & & & & \multicolumn{2}{c}{BEAM} \\
\cmidrule(lr){5-6}\cmidrule(l){10-11}
\textbf{Method} & PM-v2 & LME-S & LoCoMo & 100K & 500K & PM-v2 & LME-S & LoCoMo & 100K & 500K \\
\midrule
\multicolumn{11}{@{}l}{\emph{Context baselines}} \\
No memory & 40.0 & 6.0 & 0.1 & 26.6 & 23.3 & 30.6 & 6.8 & 0.1 & 27.2 & 25.3 \\
Full context & 49.3 & 68.0 & \underline{86.6} & 57.3 & 40.8 & 44.6 & 68.8 & \textbf{90.0} & 56.0 & 39.9 \\
Raw retrieval & 53.3 & 78.8 & 84.7 & 64.2 & 57.2 & \underline{50.6} & \underline{80.8} & 88.1 & 60.7 & 55.5 \\
\midrule
\multicolumn{11}{@{}l}{\emph{Memory systems}} \\
Mem0 \citep{chhikara2025mem0} & 45.0 & 65.6 & 57.8 & \underline{69.4} & \textbf{66.8} & 45.2 & 64.4 & 62.4 & \textbf{68.8} & 59.9 \\
MemOS \citep{li2025memos} & \underline{53.8} & 73.0 & 75.7 & 67.0 & 62.9 & 50.5 & 74.0 & 78.8 & 64.4 & \underline{61.3} \\
EverOS \citep{hu2026evermemos} & 53.6 & \underline{80.4} & 81.5 & \textbf{69.5} & 58.3 & 49.6 & 80.0 & 86.1 & \underline{67.5} & 57.0 \\
\textbf{Reglos (ours)} & \textbf{67.7} & \textbf{84.0} & \textbf{87.2} & 67.9 & \underline{65.6} & \textbf{62.1} & \textbf{87.2} & \underline{89.5} & 65.9 & \textbf{65.0} \\
\bottomrule
\end{tabular}
\caption{Results with two answering models. Qwen3.8-27B writes memory in every row. PM-v2: PersonaMem-v2; LME-S: LongMemEval-S; BEAM: mean rubric score. Best in bold, second underlined.}
\label{tab:main}
\end{table*}

\paragraph{Setting.}
We evolve the harness on PersonaMem-v2 \citep{jiang2025personamemv2}, whose four-way questions turn on implicit, revised or retracted preferences, on preferences that belong to someone else, and on information the user asked not to repeat. Qwen3.8-27B writes memory and answers throughout. The loop sees only the validation split. We report all PersonaMem-v2 results on the official 5,000-question test set.

\begin{figure}[!ht]
\centering
\includegraphics[width=\columnwidth]{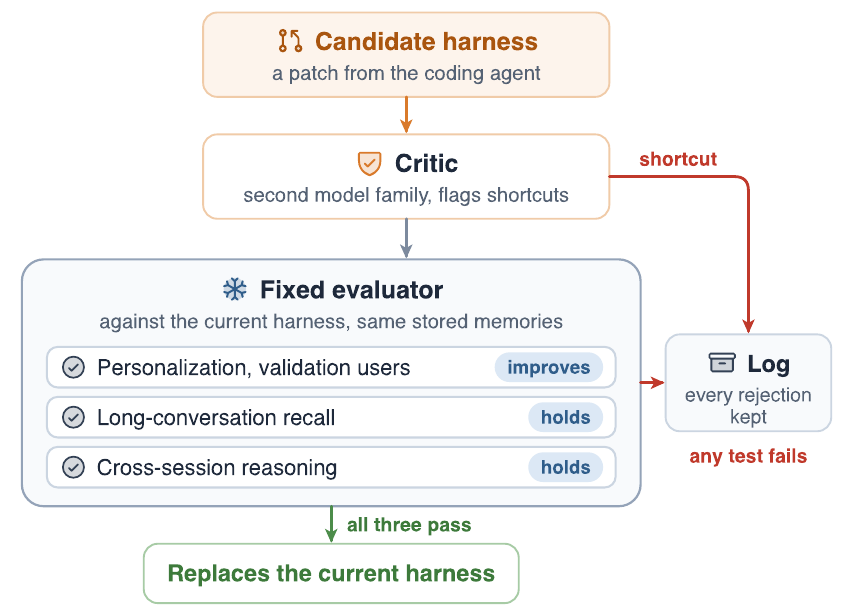}
\caption{After the freeze, a patch replaces the current harness when the critic finds it free of shortcuts, personalization improves for the validation users, and the other memory abilities hold.}
\label{fig:round}
\end{figure}

\paragraph{Proposer, critic and evaluator.}
The proposer is the coding agent Claude Code \citep{anthropic2025claudecode} running Claude Opus 5.5. It reads raw failure traces and writes a patch to the harness, following the finding that raw traces carry more signal than summaries \citep{lee2026metaharness}. The critic is Codex \citep{openai2025codex} running GPT-6.1 Sol, so the proposer and the critic come from different model families. The critic clusters the same failures independently and rejects any patch keyed to the dataset's option templates or labels. A fixed evaluator then decides. After the freeze, a patch replaces the current harness when it improves personalized answers for the validation users and preserves its other memory abilities, such as long-conversation recall and cross-session reasoning (Figure~\ref{fig:round}). The accepted harness is then scored once on the test set, whose users the loop has never seen. In the taxonomy of \citet{duan2026lastai} the outer loop is strategy autonomy (L2), and the inner loop is environment adaptation (L4); Appendix~\ref{app:run} maps both loops onto their components.

\paragraph{A trained retriever and three rules.}
\label{sec:retriever}
The loop then added a rolling preference profile to the harness. Its failure traces showed that recall driven by the request rarely reaches the exchange in which a preference was stated. The proposer therefore added a retriever to the harness: bge-base-en-v1.5, fine-tuned on 631 users of the official PersonaMem-v2 training set to rank a user's exchanges for a request. Training excludes the users of the 744 validation questions on which the patch was selected and all test users. The retriever is the only trained component, and the task model stays frozen. Three rules use it. First, the five highest-ranked exchanges follow the memory in the context as dated evidence. Second, a silence rule withholds the memory when the first-ranked exchange is tied to a request to forget or describes only another person's life, and the model answers without it. Third, when the request comes with candidate replies, a reply check compares each one with the user's requests to forget and with other people's details, and flags the replies that rely on them. The patch was then scored once on the test set (Section~\ref{sec:res-loop}); Appendix~\ref{app:run} gives the full procedure.

%% file: sections/6_results.tex
\section{Results}
\label{sec:results}

\paragraph{Models.}
Qwen3.8-27B runs with reasoning disabled through Alibaba Cloud's DashScope API, and qwen3.7-text-embedding embeds the memory. GPT-5.6 Luna answers with its default reasoning setting. When a Luna reply contains no parsable option letter, one follow-up message asks for the letter alone, which happens for 40 of the 5,000 PersonaMem-v2 questions.

\paragraph{Baselines.}
We host MemOS and EverOS ourselves, and Appendix~\ref{app:variants} describes how each is configured for PersonaMem-v2. Mem0 uses its open-source extraction prompts and retrieval scoring, and its add and search logic runs on an in-process store. On LongMemEval-S and LoCoMo (categories 1 to 4), gpt-5.5 judges each answer, and on BEAM the rubric grader of OmniMemEval \citep{omnimemeval2026} scores each answer with the same model. For BEAM-100K and BEAM-500K we sample ten conversations each, with 200 questions in total per size. The full-context baseline reads the newest 400k characters of each history.

\subsection{PersonaMem-v2}
\label{sec:res-pm}

\paragraph{Reglos is best on PersonaMem-v2 with both readers.}
Table~\ref{tab:main} compares each system in its standard configuration, with Qwen3.8-27B writing memory in every row and answering in its left half. Reglos reaches 67.7\%, 13.9 points above the strongest competitor and 18.4 above the same model reading the full history. With GPT-5.6 Luna answering, Reglos reaches 62.1\%, 11.5 points above the strongest baseline and 17.5 above the full history. A control tests whether our policy alone explains the lead. Our memory-use policy lifts every baseline by 3 to 6 points, and EverOS with the policy still trails Reglos by 8.8 (Table~\ref{tab:controls}). Published results on PersonaMem-v2 reach 57.2\%, from MemStack with a GPT-4.1-mini reader \citep{liu2026d2acci}, and 59.3\%, from DCPM with kimi-2.5 \citep{fei2026dcpm}.

\begin{table}[!ht]
\centering
\small
\setlength{\tabcolsep}{4pt}
\begin{tabular}{@{}lrr@{}}
\toprule
\textbf{Baseline with our policy} & \textbf{Accuracy} & \textbf{Reglos lead} \\
\midrule
EverOS + policy & 58.9 & $+8.8$ \\
MemOS + policy & 58.4 & $+9.3$ \\
Full context + policy & 55.6 & $+12.1$ \\
\textbf{Reglos} & \textbf{67.7} & \\
\bottomrule
\end{tabular}
\caption{Baselines given our memory-use policy, with Qwen3.8-27B answering.}
\label{tab:controls}
\end{table}

\begin{figure}[!ht]
\centering
\includegraphics[width=\columnwidth]{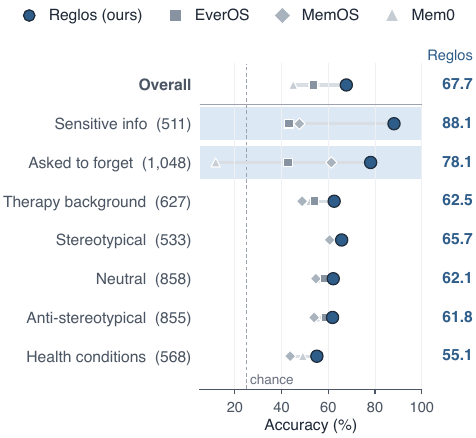}
\caption{PersonaMem-v2 accuracy by question type, with Qwen3.8-27B answering. Reglos leads most on sensitive information and forgetting.}
\label{fig:slices}
\end{figure}

\paragraph{Reglos's lead comes from sensitive information and forgetting.}
Figure~\ref{fig:slices} breaks the results down by question type. On sensitive-information questions Reglos answers 88.1\%, against 42.9\% to 47.6\% for the other memory systems. Where the user asked to forget, Reglos answers 78.1\%, ahead of MemOS (61.3\%), EverOS (42.8\%) and Mem0 (11.8\%). The tombstone and the memory-use policy target these two types, because they tell the reader what it must not use. On the five ordinary preference types Reglos is ahead on four, by 8.4 points on therapy background and by 0.4 to 3.5 points on the other three, and EverOS is ahead by 1.1 points on health conditions. The hardest questions are the 522 about a preference that belongs to someone other than the user. Reglos answers 35.4\% of them correctly, and the other memory systems answer 20.5\% or less.

\paragraph{The traced errors arise in writing memory and in using the evidence.}
To see where Reglos still fails, we took 159 questions it answered wrongly during development and traced each one to the step that caused the error. For each question, gpt-5.5 checked whether the fact it depends on appears in the context, in the stored memory and in the conversation. We then verified the labels by hand (Table~\ref{tab:errors}). In 31\% of the errors the evidence reached the context and the task model chose another option, in 16\% extraction left a stated fact out of memory, and in 4\% recall missed a stored fact. The gains left are in writing the facts that extraction misses and in helping the model use the evidence it already has, which is where the outer loop makes its changes (Section~\ref{sec:res-loop}).

\begin{table}[!ht]
\centering
\small
\begin{tabular}{@{}lr@{}}
\toprule
\textbf{Cause of the wrong answer} (159 traced) & \textbf{Share} \\
\midrule
Fact never stated in the conversation & 48\% \\
Evidence in context, reader chose wrongly & 31\% \\
Stated but never written to memory & 16\% \\
Written but not recalled & 4\% \\
\bottomrule
\end{tabular}
\caption{Causes of 159 wrong answers traced during development.}
\label{tab:errors}
\end{table}

\subsection{Generalization Across Benchmarks}
\label{sec:res-other}

\paragraph{Reglos leads the memory systems on LongMemEval-S and LoCoMo with both readers.}
On LongMemEval-S \citep{wu2025longmemeval} Reglos is ahead of the next system by 3.6 points with Qwen3.8-27B and by 6.4 with GPT-5.6 Luna (Table~\ref{tab:main}). On LoCoMo \citep{maharana2024locomo} Reglos exceeds EverOS by 5.7 points with Qwen3.8-27B and by 3.4 with GPT-5.6 Luna; with GPT-5.6 Luna the full history reaches 90.0\% and Reglos 89.5\%. LoCoMo histories are short enough that Reglos passes the reader the full dated history, so these numbers measure the reader and the prompt.

On BEAM \citep{tavakoli2026beam} Reglos ranks first at 500K tokens with GPT-5.6 Luna (65.0 against 61.3 for MemOS). With Qwen3.8-27B it scores 65.6 there, 1.2 points behind Mem0. At 100K tokens it ranks third with both readers, 1.6 and 2.9 points behind the leader. On a 1M-token sample of two conversations and 40 questions, Reglos scores 67.6 against 64.3 for Mem0.

\paragraph{Reglos leads on multi-session and temporal questions.}
With Qwen3.8-27B answering, Reglos ranks first on the two largest LongMemEval-S question types (Table~\ref{tab:types}): 76.7\% against 72.9\% for EverOS on multi-session questions and 84.2\% against 83.5\% on temporal ones. Its largest margin over EverOS is on what the assistant said earlier (98.2\% against 67.9\%). The extracted facts hold mostly what the user said, and Reglos keeps raw conversation excerpts beside them. Raw retrieval returns conversation rounds and reaches 100.0\% on this type. Its one clear weakness is the 30 preference questions, where EverOS and Mem0 lead by 13 to 17 points. On LoCoMo with Qwen3.8-27B, Reglos leads the memory systems in every category, by 6.2 points on single-hop and 10.0 on multi-hop questions over EverOS (Table~\ref{tab:locomo}). At BEAM-500K with Qwen3.8-27B, Reglos is ahead of Mem0 on temporal reasoning, abstention and contradiction resolution, and Mem0 is ahead on multi-session reasoning and knowledge updates.

\begin{table}[!ht]
\centering
\small
\setlength{\tabcolsep}{2.2pt}
\begin{tabular}{@{}lrrrrr@{}}
\toprule
\textbf{Type} (n) & \textbf{Reglos} & \textbf{EverOS} & \textbf{RAG} & \textbf{MemOS} & \textbf{Mem0} \\
\midrule
Multi-session (133) & \textbf{76.7} & 72.9 & 66.9 & 66.2 & 64.7 \\
Temporal (133) & \textbf{84.2} & 83.5 & 75.2 & 75.2 & 43.6 \\
Knowledge update (78) & 85.9 & 85.9 & \textbf{88.5} & 79.5 & 75.6 \\
User facts (70) & \textbf{94.3} & \textbf{94.3} & 90.0 & 88.6 & 88.6 \\
Assistant said (56) & 98.2 & 67.9 & \textbf{100.0} & 60.7 & 73.2 \\
Preferences (30) & 60.0 & \textbf{76.7} & 56.7 & 63.3 & 73.3 \\
\bottomrule
\end{tabular}
\caption{LongMemEval-S accuracy by question type, with Qwen3.8-27B answering. RAG: raw retrieval.}
\label{tab:types}
\end{table}

\subsection{The Outer Loop}
\label{sec:res-loop}

\paragraph{The coding agent lifted the harness by 6.5 points.}
The loop's first accepted change was an engine patch, which changed what the kernel stores, including how it records a request to forget. After it, four accepted steps took the harness from 54.2\% to 60.7\% on the 1,317 validation questions, overtaking the full history given the same policy (55.0\%) at its first step (Figure~\ref{fig:curve}). Table~\ref{tab:trajectory} in Appendix~\ref{app:run} describes each step; the largest single gain came from a revised policy that treats memory as incomplete and keeps sensitive identifiers private.

\begin{figure}[!ht]
\centering
\includegraphics[width=\columnwidth]{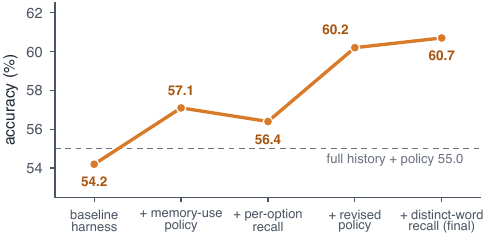}
\caption{The coding agent lifts the harness from 54.2\% to 60.7\% in four accepted steps on 1,317 validation questions. Dashed: full history with the same policy.}
\label{fig:curve}
\end{figure}

\paragraph{Forgetting as retraction.} In the engine before the patch a message such as ``please forget that I take sertraline'' was stored as an emphasized memory, because the extraction prompt read it as a request to remember. On questions that test forgetting the correct option is the generic one, and it answered none of nine development cases. Retraction with a tombstone, and a policy line that forbids relying on anything listed as forgotten, raised this to eight of nine on development and twelve of fourteen on a separate slice.

\paragraph{Evidence the reader never saw.}
Many preferences sat in raw conversation excerpts without being extracted, and recall driven by the question could not reach them, because the question does not name the preference. Recall run once per answer option, with the words that distinguish that option, finds them. Together with the policy revision between them, the four harness steps raised accuracy from 54\% to 68\% on a separate 50-question slice.

\begin{figure}[b]
\centering
\includegraphics[width=\columnwidth]{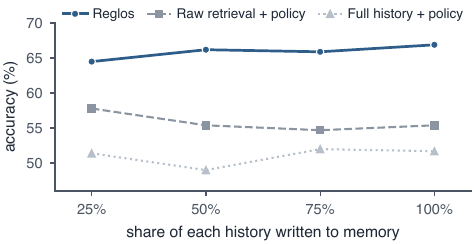}
\caption{Accuracy on early-history questions as more of each history is written (296 questions, 40 personas, Qwen3.8-27B answering).}
\label{fig:a8}
\end{figure}

\paragraph{The trained retriever and its three rules add 4.3 points on the test set.}
With the retriever and its three rules (Section~\ref{sec:retriever}), the harness scores 67.7\% on the test set against 63.4\% for the harness it started from, and 69.6\% against 64.5\% on 744 validation questions. The gain comes from questions that test forgetting (64.4\% to 78.1\% on the test set) and another person's details (21.3\% to 35.4\%). Own-preference questions are level (65.6\% and 66.1\%), and private-data questions fall from 91.8\% to 88.1\%. With one part removed at a time, validation accuracy is 63.4\% without the reply check, 65.9\% without the silence rule, 66.7\% without the retrieved exchanges, and 48.3\% with the untrained retriever. The patch changes the context only for requests that ask for help with a task. On those requests it stays level with its starting point on BEAM-1M (88.1 against 86.7 on 50 questions) and on LongMemEval-S (69.4 for both on 36 questions). Answering takes 4.5 more model calls per question.

\paragraph{What it costs.}
Writing one persona's 32k-token history to memory costs about 225k to 260k task-model tokens (roughly 7.5 times the history) and 40k embedding tokens, amortized over 25 questions per persona. Answering costs about 8.3k prompt tokens per question, so scoring one candidate on the 744 validation questions takes about 6.2M prompt tokens. Writing memory for the 200 test personas took about 50M task-model tokens and six hours at eight-way concurrency. At write time the harness also makes two short calls for each of their 25,442 exchanges, one that writes a typed note and one that labels whose life the exchange describes.

\subsection{Memory over a Growing History}
\label{sec:res-a8}

\paragraph{Accuracy holds as more is written.}
To test whether memory degrades as a history grows, we wrote only the first 25, 50, 75 or 100\% of each history for 40 personas of the benchmark split and asked the 296 questions whose evidence lies in the first quarter (Figure~\ref{fig:a8}). Reglos stays between 64.5\% at the first quarter and 66.9\% with the full history, and leads at every prefix, by 6.7 to 11.5 points over raw retrieval and by 13.1 to 17.2 points over the full history, both with our policy.

%% file: sections/7_discussion.tex
\section{Discussion}
\label{sec:discussion}

\paragraph{What the loop found.}
Each accepted change, from how a request to forget is recorded to what recall asks for, began as a failure trace that the coding agent read and turned into code.

\paragraph{What keeps the two loops trustworthy.}
Both loops rely on practices that systems software uses to run unreliable programs safely. In the inner loop the record gives recoverable state and provenance: no fact is overwritten, each one points to its source messages, and the user can undo any memory (Section~\ref{sec:inner}), while recall scans only that user's rows (Section~\ref{sec:engine}). In the outer loop the harness reaches memory through a narrow interface and never sees the labels (Section~\ref{sec:harness}). A patch replaces the current harness when it improves on the validation questions and holds on two other benchmarks, and the latest one carries its gain over to test users the loop never saw (Section~\ref{sec:res-loop}). The log keeps every candidate with its result.

\paragraph{The hardest slice.}
Preferences that belong to other people remain hard because a preference stated about a sister or a colleague is extracted as the user's, and every later layer trusts it. The fix belongs in the write path, with provenance that survives consolidation.

\paragraph{Limitations.}
One open 27B model writes memory for every system; Reglos remains first on PersonaMem-v2 and LongMemEval-S when GPT-5.6 Luna answers. The published numbers we compare with use different readers (GPT-4.1-mini and kimi-2.5). The LoCoMo result rests on reading short histories in full, and on BEAM the systems sit within a few points of each other.

\paragraph{Next steps.}
The next step in the taxonomy of \citet{duan2026lastai} is a loop that chooses which experience to gather. The product already records when users correct, delete or undo a memory; turning those events into the failure traces the proposer reads, behind the same evaluator, is the step we are taking next \citep{liang2026pahf,xue2026pastbench}. A second direction is to stop holding the model fixed. Everything here was gained with a frozen model, so training the model that writes and reads memory is the complementary step. Reinforcement learning on memory construction and use is an active direction \citep{wang2025memalpha,yan2025memoryr1}, an RL-trained 4B agentic memory reaches 55.2\% on PersonaMem-v2 \citep{jiang2025personamemv2}, and harness search can alternate with training on the policy \citep{wei2026selfhpo}. Further on, the record's operations (add, supersede, retract, consolidate, recall) could become the training objective of a memory foundation model. A third direction is modality: in our benchmark runs images enter memory only as captions, and the record, recall and redaction should extend to images, voice and the screen, under the same rules for supersession and forgetting.

%% file: sections/8_conclusion.tex
\section{Conclusion}
\label{sec:conclusion}

We presented Reglos, a personal memory agent whose two loops co-evolve at two time scales: each user's memory evolves online in an append-only record, and the harness that turns memory into context evolves offline through patches that a coding agent proposes and a fixed evaluator scores. With one frozen open 27B model, Reglos outperforms the memory systems we compared on PersonaMem-v2, LongMemEval-S and LoCoMo with either answering model, and exceeds the best published result on PersonaMem-v2. The largest share of the memory-related errors we traced are decisions the reader makes with the evidence already in view. The outer loop has so far learned from benchmark failures. Our next step is to let it learn from users directly, treating each memory a user corrects or deletes as a failure to fix.

%% file: sections/A_impl.tex
\section{Implementation Details}
\label{app:impl}

\paragraph{Recall settings.}
Recall fuses three content routes (dense vector search, keyword search and traversal of the entity graph) and two anchor routes (a time window, and a scene or location) with weighted reciprocal rank fusion ($k=60$). It keeps the 45 highest-ranked items and packs them in rank order into 8,100 tokens; both limits widen for questions that ask to enumerate.

\paragraph{The memory-use policy.}
The policy is the opening block of the system message that carries the recalled memory (Section~\ref{sec:harness}). The model receives it verbatim:

\begin{promptbox}{Prompt: memory-use policy}
\textbf{How to use this memory}
\begin{itemize}[leftmargin=1em,itemsep=1pt,topsep=1pt]
\item Personalize with the user's OWN, current preferences; prefer the response grounded in a specific remembered detail about this user.
\item This memory is incomplete. A response that mentions a plausible detail about the user that the memory does not contain is not wrong for that reason alone; reject a response only when the memory contradicts it.
\item Never rely on anything the user asked to forget (listed below). A response personalized around such a detail is wrong; prefer one that does not use it, even if it is more generic.
\item If a preference is marked as no longer true, do not use the old one.
\item Preferences of other people (friends, family, colleagues, recipients) are not the user's own.
\item Do not assume preferences from demographics or stereotypes.
\item Never repeat sensitive identifiers back to the user (passport or ID numbers, account or card numbers, email addresses, phone numbers, home addresses, file paths or project names that reveal a person or organization, credentials). When the request contains such data, the better response withholds or redacts it.
\end{itemize}
\end{promptbox}

%% file: sections/A_variants.tex
\section{Controls and Variants on PersonaMem-v2}
\label{app:variants}

Table~\ref{tab:variants} lists every PersonaMem-v2 run behind Table~\ref{tab:main}: each baseline in its standard configuration, followed by up to three variants. A \emph{+ policy} row adds Reglos's memory-use policy (Appendix~\ref{app:impl}) to the baseline's answer prompt. MemOS and EverOS follow the PersonaMem-v2 setup of the OmniMemEval harness \citep{omnimemeval2026}: 20-message sessions and top-20 retrieval, with only the models replaced. That setup leaves out the persona description that opens each conversation, which full context, Mem0 and Reglos all receive. A \emph{+ persona card} row adds this description to the start of the answer prompt and answers from the same retrieved memories.

\begin{table}[!t]
\centering
\small
\setlength{\tabcolsep}{2.5pt}
\begin{tabular}{@{}lrr@{}}
\toprule
\textbf{System} & \textbf{Accuracy} & \textbf{Reglos lead} \\
\midrule
No memory & 40.0 & $+27.7$ \\
\quad + policy & 43.1 & $+24.6$ \\
Full context (official) & 49.3 & $+18.4$ \\
\quad + policy & 55.6 & $+12.1$ \\
Raw retrieval & 53.3 & $+14.4$ \\
\quad + policy & 58.0 & $+9.7$ \\
Mem0 & 45.0 & $+22.7$ \\
\quad + policy & 49.0 & $+18.7$ \\
MemOS (OmniMemEval) & 53.8 & $+13.9$ \\
\quad + policy & 58.4 & $+9.3$ \\
\quad + persona card & 52.3 & $+15.4$ \\
\quad + policy + persona card & 57.2 & $+10.5$ \\
EverOS (OmniMemEval) & 53.6 & $+14.1$ \\
\quad + policy & 58.9 & $+8.8$ \\
\quad + persona card & 52.6 & $+15.1$ \\
\quad + policy + persona card & 58.4 & $+9.3$ \\
\textbf{Reglos} & \textbf{67.7} & \\
\bottomrule
\end{tabular}
\caption{Every control and variant on PersonaMem-v2, with Qwen3.8-27B answering.}
\label{tab:variants}
\end{table}

\begin{table}[!t]
\centering
\small
\setlength{\tabcolsep}{3pt}
\begin{tabular}{@{}lrrrr@{}}
\toprule
\textbf{Category} (n) & \textbf{Reglos} & \textbf{EverOS} & \textbf{MemOS} & \textbf{Mem0} \\
\midrule
Single-hop (841) & \textbf{92.9} & 86.7 & 81.3 & 74.2 \\
Multi-hop (282) & \textbf{83.0} & 73.0 & 69.1 & 72.7 \\
Temporal (321) & \textbf{83.2} & 82.9 & 72.6 & 5.0 \\
Open-domain (96) & \textbf{63.5} & 56.2 & 56.2 & 46.9 \\
\bottomrule
\end{tabular}
\caption{LoCoMo accuracy by category, same runs as Table~\ref{tab:main}. Mem0 stores no dates, hence its temporal score.}
\label{tab:locomo}
\end{table}

%% file: sections/A_run.tex
\section{Outer-Loop Details}
\label{app:run}

\paragraph{The two loops as improvement loops.} Table~\ref{tab:loops} maps both loops of Reglos onto the components and autonomy levels of the self-improvement framework of \citet{duan2026lastai}.

\begin{table}[!t]
\centering
\small
\setlength{\tabcolsep}{3pt}
\begin{tabular}{@{}p{0.18\columnwidth}p{0.37\columnwidth}p{0.40\columnwidth}@{}}
\toprule
 & \textbf{Inner loop} (online) & \textbf{Outer loop} (offline) \\
\midrule
Target & that user's record, summaries, profile & harness code and prompts; in one step, the engine's write rules \\
Experience & the user's conversations, calendar, corrections & failure traces on the selection questions \\
Improver & the fixed write pipeline and Dream & proposer (Claude Code) and critic (Codex) \\
Strategy & fixed rules: extraction, merge, retraction, Dream & cluster failures, form a hypothesis, patch, evaluate \\
Verifier & fixed write rules plus user review & fixed evaluator, one test-set run \\
Inherited & the append-only record & the accepted harness, the candidate log \\
Humans & what to share, what to correct or delete & objective, splits, acceptance rule, when to freeze \\
Level & L4, adaptation & L2, strategy autonomy \\
\bottomrule
\end{tabular}
\caption{The two loops of Reglos in the self-improvement framework.}
\label{tab:loops}
\end{table}

\paragraph{Splits and acceptance.}
During development the loop tuned on small slices of the selection questions (Table~\ref{tab:sets}) and checked progress on a separate slice at milestones. A change was kept when it did not lower accuracy on the tuning slices, and it replaced the current harness once it gained about 3 points. After the freeze a patch replaces the current harness when it improves on the selection questions and holds long-conversation recall (BEAM-1M) and cross-session reasoning (LongMemEval-S) (Figure~\ref{fig:round}): on the requests of these two benchmarks that the harness labels as tasks (50 and 36 questions), the patch scores at least as high as the current harness. The harness that the retriever patch starts from is the last step of Table~\ref{tab:trajectory} plus a rolling preference profile (62.9\% on its questions).

\begin{table}[!t]
\centering
\small
\setlength{\tabcolsep}{2.5pt}
\begin{tabular}{@{}llrrl@{}}
\toprule
\textbf{Set} & \textbf{Split} & \textbf{Personas} & \textbf{Items} & \textbf{Use} \\
\midrule
Training & train & 631 & 13,040 & retriever pairs \\
Selection & validation & 170 & 744 & traces, selection \\
Trajectory & validation & & 1,317 & Table~\ref{tab:trajectory} \\
Test & test & 200 & 5,000 & one final run \\
\bottomrule
\end{tabular}
\caption{PersonaMem-v2 sets. Training, selection and test personas are disjoint; trajectory uses other validation personas.}
\label{tab:sets}
\end{table}

\begin{table}[!t]
\centering
\small
\setlength{\tabcolsep}{3pt}
\begin{tabular}{@{}lp{0.80\columnwidth}r@{}}
\toprule
\textbf{Step} & \textbf{Change} & \textbf{Acc.} \\
\midrule
0 & engine: conversation-language storage; forget as retraction; implicit-preference rule & \\
& harness before any accepted harness step & 54.2 \\
1 & + forgotten facts in context + memory-use policy & 57.1 \\
2 & + per-option recall (full option text) & 56.4 \\
3 & + revised policy (incomplete memory; sensitive IDs) & 60.2 \\
\textbf{4} & + per-option recall on distinct words + raw excerpts & \textbf{60.7} \\
\bottomrule
\end{tabular}
\caption{Accepted steps of the outer loop on the 1,317 trajectory questions. Every row includes the engine patch.}
\label{tab:trajectory}
\end{table}

\paragraph{The trained retriever and its rules.}
Training uses 13,040 pairs of a request and its source exchange from 631 users of the official training set (Table~\ref{tab:sets}). The request is the benchmark's query field, and the positive is the exchange that contains the benchmark's source-conversation snippet for that query; the answer options and the correct answer are not used. Each example adds a random other exchange of the same user as a negative, with in-batch negatives under the multiple-negatives ranking loss. We fine-tune bge-base-en-v1.5 with sentence-transformers for three passes at batch size 32, learning rate $3\times10^{-5}$, warm-up over the first 10\% of steps and maximum sequence length 256. The request carries the model's query instruction, and an exchange is encoded as the first 1,500 characters of the user message and the first 500 of the assistant's reply.

At write time the harness makes two calls to the task model for each exchange. One writes a typed note of up to eight lines, each giving a detail with its owner, its area and whether it was stated or implied; a request to forget becomes a line marked \texttt{FORGET}. The other labels whose life the exchange describes (\texttt{SELF}, \texttt{OTHER} or \texttt{NONE}). At answer time a one-word call labels the request as recall of the past or as a task. Recall requests take the existing path unchanged, and for a task request the retriever ranks the user's exchanges. The silence rule fires when the first-ranked exchange is a request to forget or is directly followed by one, or when it is labelled \texttt{OTHER} and its note has no line owned by the user; the model then reads the policy and an empty memory. Otherwise the five highest-ranked exchanges follow the memory verbatim with their dates and notes, leaving out any exchange that restates a forgotten item, and the forget list follows them. With candidate replies, the reply check runs on both paths. One call compares the candidates with the forget list and with the details the notes assign to other people and names the items a candidate may rely on, and a one-word call then confirms each pair of item and candidate. The context shows each confirmed reply with its item, marked as forgotten or as another person's detail, and stays unchanged when the check confirms none or all of them.